%% file: main.tex
\documentclass[11pt]{article}

\usepackage[preprint]{acl}
\usepackage{times}
\usepackage{latexsym}
\usepackage[T1]{fontenc}
\usepackage[utf8]{inputenc}
\usepackage{microtype}
\usepackage{graphicx}
\usepackage{xurl} %

\usepackage{booktabs}
\usepackage{multirow}
\usepackage{amsmath}
\usepackage{amssymb}
\usepackage{xcolor}
\usepackage{xspace}
\usepackage{enumitem}
\usepackage{pifont}
\newcommand{\cmark}{\ding{51}}
\newcommand{\xmark}{\textcolor{gray}{\ding{55}}}
\usepackage{listings}
\usepackage[most]{tcolorbox}
\IfFileExists{fontawesome5.sty}{\usepackage{fontawesome5}}{%
  \newcommand{\faPenNib}{\raisebox{0.5pt}{\scriptsize$\blacktriangleright$}}%
  \newcommand{\faSearch}{\raisebox{0.5pt}{\scriptsize$\blacktriangleright$}}%
  \newcommand{\faHighlighter}{\raisebox{0.5pt}{\scriptsize$\blacktriangleright$}}%
  \newcommand{\faLightbulb}{\raisebox{0.5pt}{\scriptsize$\blacktriangleright$}}}
\definecolor{PPLXTurquoise}{HTML}{20808D}
\definecolor{PPLXDeep}{HTML}{0B5A63}
\definecolor{PPLXTint}{HTML}{F2F8F7}
\tcbset{fancyprompt/.style={
  enhanced, breakable,
  colback=PPLXTurquoise!4!white, colframe=PPLXTurquoise!55!white,
  boxrule=0.5pt, arc=2.6mm,
  left=8pt, right=8pt, top=11pt, bottom=8pt,
  fontupper=\normalsize,
  colbacktitle=PPLXTurquoise!85!white, coltitle=white,
  fonttitle=\bfseries\normalsize,
  attach boxed title to top left={yshift=-2.6mm, xshift=3.5mm},
  boxed title style={arc=1.4mm, boxrule=0pt, left=6pt, right=6pt, top=2.6pt, bottom=2.6pt},
  fuzzy shadow={0mm}{-0.5mm}{0mm}{0.2mm}{PPLXDeep!12!white}}}

\newcommand{\dataname}{\textsc{PII-TRACE}\xspace}
\newcommand{\modelname}{PII-Tracer\xspace}

\title{\dataname: A Benchmark for Context-Aware PII Detection\\ in Multi-Turn LLM Conversations}

\author{Kaiyuan Zhang\textsuperscript{1,2*}, 
Chuan Wang\textsuperscript{1*},
Joey Zhong\textsuperscript{1}, 
Paul Fryzel\textsuperscript{1},\\\bf Kyle Polley\textsuperscript{1}, Jerry Ma\textsuperscript{1}, \and Ninghui Li\textsuperscript{1,3}, \\
  \textsuperscript{1}Perplexity, 
  \textsuperscript{2}Rutgers University, 
  \textsuperscript{3}Purdue University\\
}

\begin{document}
\maketitle
\begingroup
\renewcommand{\thefootnote}{*}
\footnotetext{Equal contribution.}
\endgroup

\begin{abstract}
LLM assistants and agentic systems log long multi-turn conversations.
AI providers often scan these conversations for Personally Identifiable Information (PII) and mask the PII before storing or processing conversation data.
Yet most PII detectors and benchmarks target self-contained records rather than cross-turn evaluation.
To evaluate PII detection across turns in multi-turn conversations, we introduce \dataname (\textbf{T}racing \textbf{R}ecurring PII \textbf{A}cross \textbf{C}onversational \textbf{E}xchanges), to our knowledge the first PII benchmark to assess whether detectors identify PII in conversational contexts and cover every mention of a recurring identifier across turns.
\dataname contains 13{,}148 synthetic multi-turn dialogues in 13 languages with character-level spans and identifier clusters.
Across eleven baselines, including frontier LLMs, no detector achieves full entity-level coverage without substantial false positives on PII-free conversations, and single-pass reading loses a third of the gold characters on long dialogues.
To close this gap, we introduce \modelname, a compact 0.6B-parameter detector trained with conversation-level supervision.
\modelname attains the highest entity-level coverage of any system we evaluate and also performs strongly on standard single-record benchmarks.
We will release both the benchmark and detector upon publication.
\end{abstract}

\input{sections/intro}

\input{sections/related}

\input{sections/benchmark}

\input{sections/experiments}

\input{sections/conclusion}

\input{sections/limitations}

\input{sections/ethics}

\bibliography{reference}

\appendix

\input{sections/appendix}

\end{document}

%% file: sections/intro.tex
\section{Introduction}
\label{sec:intro}
\input{figures/moti_example}

Scanning text for Personally Identifiable Information (PII) is a standard component of data pipelines in modern LLM systems. Providers detect and remove PII when curating pretraining corpora \citep{dolma2024, roots2022, starcoder2023, llama3herd2024} and before storing or reusing collected data. This process helps satisfy data protection regulations \citep{gdpr2016} and reduces the risk that models memorize and later reveal sensitive information \citep{carlini2021extracting}. Increasingly, PII detectors are built on transformer-based language models, including a recently released open-weight bidirectional token classifier \citep{openaiprivacyfilter2026}.

Existing PII detectors are typically designed for relatively short, self-contained records rather than the long, multi-turn conversations common in LLM assistants and agentic systems. Within a single conversation, users may paste emails or medical reports, switch between languages, and interact with assistants that call external tools or share context with other agents \citep{yao2023react, xi2023agents, packer2024memgpt}. Detectors designed to process isolated records may therefore struggle with the length, complexity, and contextual dependencies of real-world conversations.

\textbf{Conversational context determines what counts as PII and which mentions must be detected.} The same text span can be PII in one conversation but not in another. For example, a name is PII when it refers to the user, but not when it refers to a public figure or a place named after that person. Moreover, the same identifier may recur throughout a conversation. In Figure~\ref{fig:teaser}, for example, the name \emph{Maria Torres} appears across multiple turns. A detector must identify every occurrence of a recurring identifier, as missing even one leaves that occurrence unprotected. This requirement becomes especially important when conversations are passed to external tools or other agents, where private content may be further exposed \citep{privacyaction2025, agentleak2026}. PII detection in conversations is therefore inherently a \textbf{context-aware} task.

However, established PII benchmarks are primarily document- or record-oriented rather than conversational \citep{ai4privacy2023, pilan2022tab, stubbs2015i2b2}. In our experiments, we identify three properties that these benchmarks do not adequately evaluate. The first is \emph{cross-turn consistency}: every mention of a recurring identifier should be detected. The second is avoiding \emph{long-context degradation}, where recall decreases as conversation length grows \citep{liu2024lost, openaiprivacyfilter2026}. The third is the ability to handle \emph{multilingual mixed-medium text}, where languages may switch within a dialogue and prose may be interleaved with code and structured records. As a result, strong performance on existing benchmarks does not necessarily translate into effective PII detection in real-world conversational settings.
The closest concurrent efforts, such as REDACT~\citep{vats2026redact} and RedactionBench~\citep{redactionbench2026}, cover more languages and longer documents, but do not provide explicit cross-turn mention chains needed to evaluate cross-turn consistency.

This paper asks: \emph{Do PII detectors consistently cover recurring identifiers in multi-turn LLM conversations?}

To answer this question, we introduce \dataname, to our knowledge the first PII benchmark with an explicit cross-turn detection task for multi-turn conversations. \dataname comprises 13{,}148 synthetic dialogues derived from production assistant traffic, with character-level annotations spanning nine identifier types. Mentions of the same identifier are further linked into entities, allowing us to evaluate whether a detector follows an identifier across the entire conversation. We call this property \emph{consistent detection}: an entity is considered covered only if every mention of that entity is detected.
Conversations in \dataname range from fewer than 1{,}000 to more than 100{,}000 characters, enabling us to measure robustness to long-context degradation. The benchmark also includes conversations that switch languages and interleave prose with code, tables, and structured records, spanning 13 languages in total. It therefore enables evaluation of detectors on multilingual mixed-medium text. Finally, \dataname includes PII-free conversations for measuring false positives.

We find that existing detectors struggle to consistently cover recurring identifiers across conversations. Across eleven baselines, ranging from open-source detectors to frontier LLMs prompted for PII detection, no public detector achieves strong consistent detection without flagging most PII-free conversations. Frontier LLMs are substantially more precise, but still miss many PII mentions.

To close this gap, we introduce \modelname, a compact 0.6B-parameter detector trained on multilingual conversations. \modelname labels an entire 4{,}096-token window in a single pass. Longer conversations can be processed using overlapping sliding windows, which substantially improves recall on long conversations (\S\ref{sec:experiments}). \modelname achieves the highest consistent detection and character-level F1 among all systems we evaluate, while using only a small fraction of the parameters of frontier LLMs (\S\ref{sec:model}, \S\ref{sec:experiments}).

\paragraph{Contributions.}
\begin{itemize}[topsep=2pt,itemsep=1pt,leftmargin=*]
\item We introduce \dataname, a multi-turn PII benchmark with an explicit cross-turn detection task. It includes long, multilingual, and mixed-medium conversations, with mentions linked into entities for evaluating \emph{consistent detection} across turns.

\item We evaluate eleven PII detectors, ranging from open-source specialized models to frontier LLMs. Public specialized detectors achieve high recall but low precision, while frontier LLMs are more precise but miss many mentions. For most detectors, recall also degrades substantially as conversations grow longer.

\item We introduce \modelname, a compact 0.6B-parameter PII detector trained on conversational data. It achieves better coverage of recurring identifiers than all other evaluated detectors, including much larger frontier LLMs.
\end{itemize}

%% file: figures/moti_example.tex
\begin{figure}[t]
\centering
\includegraphics[width=\columnwidth]{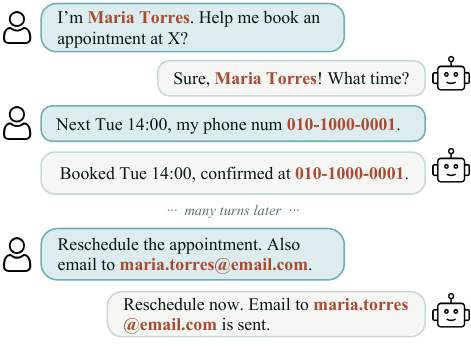}
\caption{One identifier (highlighted) recurs across a conversation. \dataname's \emph{consistent-detection} metric counts it as covered only if every mention is found.}
\label{fig:teaser}
\end{figure}

%% file: sections/related.tex
\input{tables/table1_compare}

\section{Related Work}
\label{sec:related}

\textbf{PII benchmarks.} Existing benchmarks cover synthetic records, including ai4privacy, Nemotron-PII, and SPY \citep{ai4privacy2023,nemotronpii2025,savkin2025spy}; legal documents with within-document coreference, such as TAB \citep{pilan2022tab}; longitudinal but non-conversational clinical records, such as i2b2/UTHealth \citep{stubbs2015i2b2}; and newswire named-entity recognition (NER), such as CoNLL-2003 \citep{tjongkimsang2003conll}. Concurrent work includes REDACT, which covers mixed formats and multi-turn chats without cross-turn mention chains \citep{vats2026redact}; the document-oriented RedactionBench \citep{redactionbench2026}; and the query-focused PII-Bench and CAPID \citep{piibench2025,capid2026}. None provides explicit mention chains across turns of a longitudinal user--assistant conversation (Table~\ref{tab:compare}).

\paragraph{PII detectors.} Existing detectors range from rule- and NER-based systems, such as Microsoft Presidio \citep{presidio}, to learned span models, including the OpenAI Privacy Filter \citep{openaiprivacyfilter2026}, GLiNER2-PII \citep{gliner2pii2026}, and open-source detectors from the Piiranha \citep{piiranha2024} and OpenMed \citep{openmedpii2026} series. We evaluate seven of these detectors on conversations in \S\ref{sec:experiments}.

\paragraph{Contextual and agentic privacy.}
Privacy is contextual: whether an information flow is appropriate depends on the context in which it occurs \citep{nissenbaum2004privacy}. LLMs can leak private information in inappropriate contexts \citep{confaide2024}, infer hidden attributes from contextual cues \citep{staab2024beyond}, and reproduce memorized training data \citep{carlini2021extracting, propile2023}. Agentic privacy benchmarks test whether agent actions respect privacy norms \citep{shao2024privacylens} and measure information leakage through tool use and inter-agent interactions \citep{privacyaction2025, agentleak2026}. Separately, long-context studies show that model performance degrades as input length increases \citep{liu2024lost, hsieh2024ruler, bai2024longbench}. \dataname is complementary to this work: it evaluates span-level PII detection across an entire conversation.

%% file: tables/table1_compare.tex
\begin{table*}[!t]
\caption{\dataname vs.\ representative PII benchmarks: only \dataname pairs multi-turn
conversational text with cross-turn identifier clusters (TAB annotates coreference within single
documents; REDACT includes flat chats among mixed formats but defines no cross-turn task). Sizes approximate.}
\label{tab:compare}
\centering
\small
\begin{tabular}{llrlcc}
\toprule
Benchmark & \#Ex. & Langs & Setting & \shortstack{Entity-cluster} & Cross-turn \\
\midrule
ai4privacy \citep{ai4privacy2023}      & 225k  & 6  & short document & \xmark & \xmark \\
Nemotron-PII \citep{nemotronpii2025}   & 100k  & 1  & document      & \xmark & \xmark \\
SPY \citep{savkin2025spy}              & 8.7k  & 1  & QA (author)   & \xmark & \xmark \\
TAB \citep{pilan2022tab}               & 1.3k  & 1  & legal doc     & \cmark & \xmark \\
REDACT \citep{vats2026redact}          & 13.4k & 25 & mixed incl. chat & \xmark & \xmark \\
PII-Bench \citep{piibench2025}         & 2.8k  & 1 & query + description & \xmark & \xmark \\
RedactionBench \citep{redactionbench2026} & 0.2k & 1 & document   & \xmark & \xmark \\
\midrule
\textbf{\dataname (ours)} & \textbf{13{,}148} & \textbf{13} & \textbf{multi-turn conv.} &
\cmark & \cmark \\
\bottomrule
\end{tabular}
\end{table*}

%% file: sections/benchmark.tex
\section{The \dataname Benchmark}
\label{sec:benchmark}

\subsection{Problem formulation}
\label{sec:task}

We study PII detection in long, multi-turn conversations. Given a complete user--assistant dialogue, the task is to identify every text span that refers to a private individual. We adopt an operational definition of PII as information that identifies a specific person, either on its own or when combined with other information. This definition draws on the GDPR \citep{gdpr2016}, the ISO/IEC 29100 privacy framework \citep{iso29100}, U.S. federal guidance \citep{nistsp800122}, and scholarship on contextual identifiability \citep{schwartz2011pii}.

Consider the name \emph{Maria Torres} in Figure~\ref{fig:teaser}. Whether this span constitutes PII cannot be determined from the string alone. If a user states their name, the span reveals personal information. 
The same string does not identify anyone when the assistant introduces it as a fictional placeholder in an example.
Distinguishing among these cases requires reasoning about the surrounding conversational context, making PII detection a \textbf{context-aware} task.

Formally, a conversation is a sequence of $m$ user and assistant turns concatenated into one text,
\begin{equation}
x \;=\; \tau_1 \,\Vert\, \tau_2 \,\Vert\, \cdots \,\Vert\, \tau_m
\end{equation}
where turn $\tau_i$ occupies the character positions $T_i$ of $x$. A detector is graded on all of $x$. It may read the text in one window or several (a choice we leave to its inference policy), but it must commit to a single set $P$ of character positions marked as PII. We do not require it to say which marked positions belong to the same entity. The task is detection only.

The nine identifier types shown in Table~\ref{tab:types} cover the label sets of deployed detectors \citep{openaiprivacyfilter2026, gliner2pii2026}.
Let $G$ be the set of gold PII characters. An \emph{entity} $E\subseteq G$ is the set of characters belonging to a single identifier, with repeated mentions grouped by coreference, and the entities $\mathcal{E}$ partition $G$.
We call an entity \emph{cross-turn} if its characters appear in more than one turn in the conversation,
\begin{equation}
\bigl|\{\, i : E \cap T_i \neq \emptyset \,\}\bigr| \ge 2 .
\end{equation}

\input{tables/table_types}

Achieving protection requires \textbf{consistent detection} because for recurring identifiers, missing even one occurrence leaves the identifier exposed.
First, at the character level, a label is considered correct if a labeled position belongs to $G$. This measures how many PII texts are detected. Second, at the entity level, an entity is considered consistently detected if all characters of entity $E$ are labeled ($E\subseteq P$). The evaluation setting (\S\ref{sec:using}) embodies these two criteria into the precise metrics reported in this paper.

\subsection{Design desiderata}
\label{sec:desiderata}
Our design goals follow the actual deployment scenario: for the LLM assistant, the PII detector deals with dialogue logs, not the short, well-structured records common in existing corpora.

Splitting a dialogue into individual records results in the loss of significant structural information, since an identifier may reappear after many rounds. We assign a cluster ID to each occurrence, enabling evaluation to track the identifier across the entire dialogue rather than a single location. Dialogue lengths vary widely, from a few hundred to over one hundred thousand characters, allowing us to measure recall based on input length. Dialogues are also not always monolingual natural language text: they may switch languages mid-conversation, and plain text may contain code, tables, or structured records. In our experiments, existing baseline detectors performed poorly primarily on dialogues with these characteristics (\S\ref{sec:experiments}).

\begin{figure*}[t]
\centering
\includegraphics[width=\textwidth]{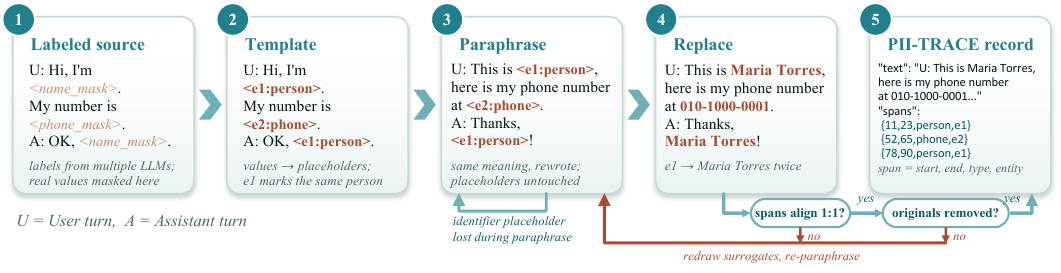}
\caption{Constructing \dataname, on one example (source values withheld): a labeled conversation
becomes a typed, cluster-preserving template; each turn is paraphrased with placeholders intact,
and each placeholder is replaced by its entity's surrogate with exact offsets. Failed paraphrases
are retried, and gate-flagged documents can re-enter synthesis with fresh surrogates before a final
drop (\S\ref{sec:verify}). Steps~1 and~2 are the anonymization of \S\ref{sec:construction}, steps~3 and~4 the
synthesis of \S\ref{sec:synthesis}, and the gates, guarding the released record (step~5), the
verification of \S\ref{sec:verify}.}
\label{fig:pipeline}
\end{figure*}

\paragraph{Pipeline.} We synthesize the benchmark from real traffic in five steps
(Figure~\ref{fig:pipeline}), labeling and anonymization (steps~1, ~2, \S\ref{sec:construction}),
synthesis (steps~3, ~4, \S\ref{sec:synthesis}), and the verification gates that decide whether a
record (step~5) is released (\S\ref{sec:verify}).

\subsection{Labeling and anonymization}
\label{sec:construction}
We begin with large-scale real-world user-assistant dialogue samples collected from a production environment, transforming each dialogue into a de-identified template. Multiple state-of-the-art LLMs are used to annotate identifiers according to the nine types of annotation in Table ~\ref{tab:types} (Figure ~\ref{fig:pipeline}, step~1). Subsequently, recurring identifiers of the same type are grouped into the same entity through rule-based processing, and sensitive attribute layers are annotated in a separate process. In step~2, we remove each identifier and replace it with a placeholder. The template retains the multi-turn structure of the dialogue, as well as the position, type, and entity ID of each occurrence, while not retaining any of the original annotated values. 

\subsection{Synthesis}
\label{sec:synthesis}
Each template in the anonymization phase generates a synthesized dialogue (Figure~\ref{fig:pipeline}, steps~3 and~4). Surrogates are generated for each entity, matching their type, format, and geographic location, and using values that are formatted correctly but not usable. The same surrogate is used for every occurrence, ensuring identifier consistency within a document, while surrogates in different documents are generated independently.
The language model then rewrites each round of dialogue (step~3) and inserts the surrogates (step~4), so both the textual representation and the annotated identifier values in the final benchmark are synthesized content. If a placeholder is lost during paraphrasing, the rewrite is regenerated up to three times (back loop in step~3); if it still fails, the document is flagged and proceeds to the verification checks in \S\ref{sec:verify}. This replacement strategy follows the clinical de-identification practice  \citep{carrell2013hips, stubbs2015i2b2}.

\subsection{Alignment, verification, and audit}
\label{sec:verify}
Every tagged identifier in the benchmark is a surrogate that the pipeline placed (Figure~\ref{fig:pipeline}, step~5), so the gold span can be accurately determined without manual annotation. The alignment step further ensures accurate positioning: after rewriting, we recalculate each surrogate's character offset and append its type and entity ID.

Subsequently, the validation process verifies the correct construction of each document through three automated checks. First, the replacement value span must correspond one-to-one with the gold mention, and all mentions of the same entity must share the same value. Second, no original values should be detected during independent rescanning using Microsoft Presidio and strict regular expressions. Third, each stored character offset must accurately extract the corresponding surrogate substring. The first two checks correspond to the decision nodes in Figure~\ref{fig:pipeline}. A document that fails any gate is resynthesized with fresh surrogates and re-paraphrased turns (the loop back into step~3), and discarded if it still fails.
Since these checks are rule-based, we also used a second, independent language model to audit a portion of the released documents as a fuzzy check; anything it flagged was manually reviewed against the document's surrogate list. See Appendix~\ref{app:prompts} for the prompts used in the pipeline.

\subsection{Composition}
\label{sec:stats}
\dataname contains 13{,}148 conversations, split into training, validation, and test sets with no source conversations shared between the different data partitions; 5,645 dialogues contain gold PIIs, and 7,503 are PII-free. 
Dialogue lengths range from less than 1,000 characters to over 100,000 characters, and thirteen languages each contain at least 100 dialogues. In dialogues containing PIIs, $63.8\%$ of the dialogues contain entities mentioned multiple times, and $28.7\%$ of the dialogues have entities that appear repeatedly across rounds; therefore, detecting only one mention often fails to cover the entire entity.

\subsection{\modelname: a context-aware detector}
\label{sec:model}
\modelname models PII detection in the dialogue as token classification. One encoder pass reads a window (\S\ref{sec:task}) of the dialogue $x$ and generates contextual states $h_{1:T}$ for $T$ tokens within it. Therefore, the annotation for each token is based on the surrounding dialogue window as context, rather than using the autoregressive prompting used in existing LLM baselines.

\paragraph{Architecture.} \modelname is a bidirectional encoder with $0.6$B parameters, using a Qwen3 \citep{qwen3_2025} backbone adapted with masked-diffusion pretraining, reading a maximum of $4096$ tokens per window. On the shared token states, the model uses a tagging head covering 37 BIOES labels (including a background tag $O$ and $\{$B, I, E, S$\}$ for each of the nine types), and an auxiliary sensitivity head used during training to predicts whether the conversation contains sensitive content as an auxiliary training signal. Gold spans are mapped to tokenization and represented as BIOES tags; therefore, detection is modeled as per-token classification with the surrounding conversation window as context.

\paragraph{Training and decoding.} We minimize
\begin{equation}
\mathcal{L}=\lambda_{\text{tok}}\,\mathcal{L}_{\text{tag}}+\lambda_{\text{sens}}\,\mathcal{L}_{\text{sens}},
\end{equation}
where $\mathcal{L}_{\text{tag}}$ is the class-weighted cross-entropy for BIOES tags, and $\mathcal{L}_{\text{sens}}$ is the binary cross-entropy for sensitivity logit, with weights of $\lambda_{\text{tok}}{=}1.5$ and $\lambda_{\text{sens}}{=}0.3$, respectively. We trained on AdamW for 3 epochs with 714k samples, including multilingual assistant conversations annotated by multiple frontier language models and samples from the ai4privacy corpus. Each training sample concatenates all turns in a conversation into plain text and annotates each mention on the conversation-level offset, enabling the model to judge each span in conjunction with the conversational context. Therefore, the same string can be supervised as a PII in one conversation and as background in another. During inference, per-token scores are decoded into typed spans while ensuring the tag sequences are valid; two learnable boundary biases adjust the tradeoff between precision and recall without retraining; for conversations longer than the window, decoding can be performed window-by-window (\S\ref{sec:experiments}). More hyperparameter settings in Appendix~\ref{app:refmodel}.

%% file: tables/table_types.tex
\begin{table}[t]
\caption{The nine identifier types, with gold mention counts over all three splits (37{,}431 total;
every occurrence counts separately).}
\label{tab:types}
\centering
\small
\resizebox{\columnwidth}{!}{%
\begin{tabular}{@{}lp{3.55cm}r@{}}
\toprule
Type & Covers & Mentions \\
\midrule
\texttt{private\_person}  & name of a private person: full name, name part, handle & 22{,}973 \\
\texttt{private\_date}    & date identifying a person & 3{,}381 \\
\texttt{private\_url}     & personal URL or IP address & 2{,}657 \\
\texttt{private\_address} & postal address, precise location & 2{,}521 \\
\texttt{account\_number}  & national ID, SSN, IBAN, card, account ids & 2{,}013 \\
\texttt{private\_email}   & personal email address & 1{,}626 \\
\texttt{private\_phone}   & personal phone or fax & 985 \\
\texttt{other\_pii}       & residual annotator-marked identifiers & 860 \\
\texttt{secret}           & credentials: passwords, keys, PINs & 415 \\
\bottomrule
\end{tabular}}
\end{table}

%% file: sections/experiments.tex
\section{Experiments}
\label{sec:experiments}

\input{tables/table4_cd}

\subsection{Evaluation setup}
\label{sec:using}
Our evaluation covers twelve detectors: Microsoft Presidio \citep{presidio}, six learned span models (the OpenAI Privacy Filter with 1.5B parameters \citep{openaiprivacyfilter2026}, GLiNER2-PII \citep{gliner2pii2026}, zero-shot GLiNER-PII \citep{nvidiaglinerpii2025}, Piiranha \citep{piiranha2024}, and two OpenMed clinical PII models \citep{openmedpii2026}), four frontier LLMs prompted zero-shot with the nine-type taxonomy (GPT-5.4 \citep{gpt54card2026}, GPT-5.6-sol \citep{gpt56card2026}, Claude Opus 4.8 \citep{opus48card2026}, and Claude Sonnet 5 \citep{sonnet5card2026}), and ours \modelname. 
We evaluate on the test split (1{,}922 documents) and report metrics as following:
\begin{itemize}[topsep=2pt,itemsep=1pt,leftmargin=*]

\item \textbf{Character P / R / F1}: A predicted character is considered a true positive if it lies within a gold span. Character F1 is our primary detection metric, measuring how many PII texts are successfully masked without considering span boundaries.

\item \textbf{Span-Overlap and Span-Containment P / R / F1}: For overlap, a prediction is considered correct if it overlaps with any gold span. For containment, precision counts predicted spans contained within a gold span, while recall counts gold spans contained within a predicted span.

\item \textbf{Consistent detection}: The percentage of gold entities whose mentions are covered. CD-multi is calculated only for multi-mention entities, while cross-turn CD is calculated only for entities that repeat across rounds.

\item \textbf{Has-PII accuracy and FP$_0$}: The former is the document-by-document binary has-PII accuracy, and the latter is the false-positive rate on documents without PII.

\end{itemize}

\input{tables/table_cdsize}

\subsection{Main results}
Table ~\ref{tab:cd} divides the twelve systems into two groups. Publicly available specialized baselines achieve high recall through over-marking. OpenMed models can cover more than $0.9$ of gold characters, but at the cost of marking more than six times the gold character volume, resulting in character precision below $0.15$. Presidio marks 16 times the gold volume, with a precision of only $0.045$, primarily covering structured identifiers such as email and phone numbers, with less coverage of free text. Frontier general-purpose LLMs exhibit the opposite pattern. They have the highest character precision, approaching $0.5$, while all publicly available specialized baselines do not exceed $0.36$; however, they only cover gold characters from $0.56$ to $0.68$.

\modelname combines the advantages of both. It can cover 0.830 gold characters, comparable to over-marking detectors, while achieving a precision of 0.507, comparable to frontier LLMs; its char-F1 is 0.629, the highest of all systems, making it the most balanced among the twelve detectors. The much larger parameter-scale GPT-5.6-sol only slightly surpasses it in span-level F1 (overlap 0.632 vs. 0.621, containment 0.612 vs. 0.580).

\subsection{Cross-turn consistency}
\label{sec:crossturn}
Covering characters is easier than covering every mention of an entity, since missing any mention leaves the entity exposed; multi-mention consistent detection is therefore lower than character recall for all detectors. \modelname covers $0.830$ of gold characters, with a multi-mention consistent detection (CD-multi) of $0.794$, the highest of all systems, significantly higher than GPT-5.6-sol ($0.570$) and other frontier LLMs ($0.24$-$0.31$). Aggregate CD, CD-multi, cross-turn CD, and PII-free false-positive rates for the twelve detectors are shown in Appendix~\ref{app:results} (Table~\ref{tab:agg}), while Table~\ref{tab:cdsize} analyzes consistent detection by mention count ($899$ single-mention and $959$ multi-mention entities, $790$ of the latter appearing repeatedly across rounds). As the number of mentions increases, consistent detection declines sharply for learned detectors (GLiNER2-PII from $0.641$ for a single mention to $0.073$ for 6–10 mentions) and frontier LLMs (GPT-5.6-sol from $0.788$ to $0.464$, Claude Opus 4.8 from $0.840$ to $0.045$), whereas \modelname declines more gradually (from $0.917$ to $0.691$) and performs best across all groups, including cross-turn entities ($0.776$). Presidio consistently performs at around $0.63$ across all groups because it labels most structured values rather than selectively covering a particular entity.

\subsection{Document-level flagging}
\label{sec:haspii}
In practice, Guardrail must also determine whether a conversation contains PII, as incorrectly labeling clean conversations incurs costs. Has-PII accuracy reveals over-flagging issues: Presidio and OpenMed models score between 0.43 and 0.45, below the Has-PII baseline because they label PII in almost every conversation; frontier LLMs score between 0.76 and 0.83; and \modelname reaches $0.759$, the highest among specialized detectors and close to frontier LLMs. The complementary metric FP$_0$ is shown in Table~\ref{tab:agg}: all public specialized detectors label more than half of the PII-free documents, while the proportion for \modelname is $0.385$. This reflects the cost of over-marking at the document level, as predicted spans scattered throughout clean conversations become false flags.

\input{tables/table_length}

\subsection{Single-window degradation}
As dialogue length increases, recall decreases. In a single 4096-token window, \modelname covers $0.975$ of gold characters in dialogues shorter than 1k characters, $0.955$ in dialogues between 1 and 10k characters, and drops to $0.687$ for dialogues longer than 10k characters (Table~\ref{tab:length}). Precision does not decrease with length; it is lowest in the shortest range ($0.392$), and around $0.51$ in other ranges. This is primarily due to single-window truncation: using 50\%-overlap sliding windows decoding for the same checkpoint can improve character recall from $0.83$ to $0.97$ and multi-mention CD from $0.79$ to $0.95$ (Appendix~\ref{app:results}). Longer dialogues contain a larger absolute number of identifiers, so documents with decreased recall face a greater risk of missed detections.
The sliding-window method incurs a slight loss of precision, but the two strategies can be chosen at decode time, so a suitable operating point can be selected during deployment without retraining.

\begin{figure}[t]
\centering
\includegraphics[width=0.98\columnwidth]{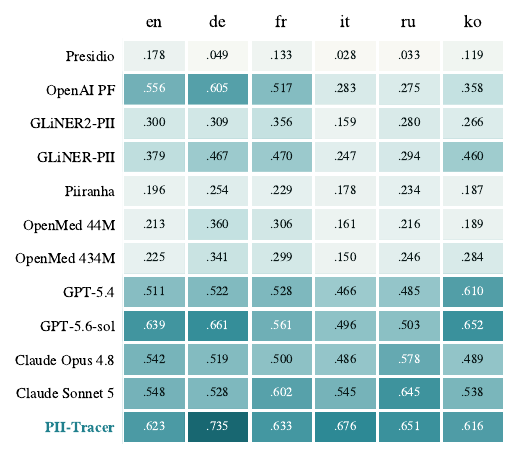}
\caption{Character F1 by language for six test languages covering the Latin, Cyrillic, and Hangul
scripts.}
\label{fig:lang}
\end{figure}

\subsection{Multilingual mixed-medium text}
Figure ~\ref{fig:lang} shows the character F1 scores for six language subsets in the corpus, covering Latin, Cyrillic, and Hangul scripts; full language-specific results will be available upon release. Each language was evaluated on all test documents, ranging from $928$ for English to $61$ for Italian. System rankings varied across different scripts. Presidio covered $0.79$ of gold characters on English but only $0.42$ on Russian; OpenMed 434M's recall dropped from $0.94$ for English to $0.69$ for Korean; OpenAI Privacy Filter maintained recall across different scripts, thus its F1 score dropped to $0.28$ on Italian and Russian, primarily due to precision.
\modelname achieved the highest character F1 score in four of the six languages (all between $0.616$ and $0.735$; GPT-5.6-sol scored slightly higher in `en' and `ko'), and the highest consistent detection across all six languages, ranging from $0.80$ to $0.93$. Because rankings vary across different scripts, selecting a detector solely based on overall or English scores may not suit for real-world language combinations.

\begin{figure}[t]
\centering
\includegraphics[width=0.98\columnwidth]{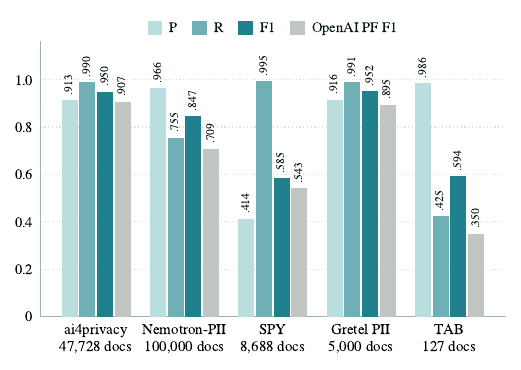}
\caption{\modelname on five external PII benchmarks (label-agnostic character P/R/F1); gray: the
OpenAI Privacy Filter under identical evaluation.}
\label{fig:external}
\end{figure}

\subsection{External benchmarks}
Figure~\ref{fig:external} shows the results of \modelname on five external PII benchmarks, including two public PII datasets used in the OpenAI Privacy Filter model card: ai4privacy and SPY.
\modelname outperforms the Privacy Filter on all benchmarks: char F1 $0.950$ vs $0.907$ on ai4privacy \citep{ai4privacy2023}, $0.847$ vs $0.709$ on Nemotron-PII \citep{nemotronpii2025}, $0.585$ vs $0.543$ on SPY \citep{savkin2025spy}, $0.952$ vs $0.895$ on the Gretel PII masking set \citep{gretelpii2024}, and $0.594$ vs $0.350$ on TAB \citep{pilan2022tab}.
TAB is the only benchmark that includes real human-annotated text, and at the same precision, \modelname achieves twice the recall. The improvement does not stem from a shared synthesis style.
Table~\ref{tab:cd} indicates that single-record detectors perform poorly in dialogue scenarios, whereas \modelname is competitive in both settings. Since its training mixture includes single-record data, we do not attribute this result entirely to conversational supervision.

%% file: tables/table4_cd.tex
\begin{table*}[!t]
\caption{PII detection on the \dataname test set (1{,}922 documents); all metrics label-agnostic,
higher is better. The four frontier LLMs are prompted zero-shot; span scores match gold spans by
overlap and by containment.}
\label{tab:cd}
\centering
\small
\setlength{\tabcolsep}{4.4pt}
\begin{tabular}{lccccccccc}
\toprule
& \multicolumn{3}{c}{Character} & \multicolumn{3}{c}{Span-Overlap} & \multicolumn{3}{c}{Span-Containment} \\
\cmidrule(lr){2-4}\cmidrule(lr){5-7}\cmidrule(lr){8-10}
Detector & P & R & F1 & P & R & F1 & P & R & F1 \\
\midrule
Presidio (rule/NER) \citep{presidio} & 0.045 & 0.762 & 0.086 & 0.042 & 0.844 & 0.079 & 0.036 & 0.741 & 0.069 \\
OpenAI Privacy Filter (1.5B) \citep{openaiprivacyfilter2026} & 0.358 & 0.785 & 0.492 & 0.266 & 0.817 & 0.401 & 0.380 & 0.363 & 0.371 \\
GLiNER2-PII \citep{gliner2pii2026}   & 0.211 & 0.475 & 0.292 & 0.181 & 0.525 & 0.269 & 0.173 & 0.514 & 0.259 \\
GLiNER-PII (zero-shot) \citep{nvidiaglinerpii2025} & 0.261 & 0.785 & 0.392 & 0.209 & 0.890 & 0.338 & 0.272 & 0.483 & 0.348 \\
Piiranha v1 \citep{piiranha2024}     & 0.161 & 0.299 & 0.209 & 0.117 & 0.339 & 0.174 & 0.078 & 0.157 & 0.104 \\
OpenMed PII (44M) \citep{openmedpii2026} & 0.136 & 0.901 & 0.236 & 0.094 & 0.942 & 0.170 & 0.040 & 0.786 & 0.077 \\
OpenMed PII (434M) \citep{openmedpii2026} & 0.145 & 0.919 & 0.251 & 0.109 & 0.957 & 0.195 & 0.047 & 0.787 & 0.089 \\
GPT-5.4 \citep{gpt54card2026}        & 0.466 & 0.569 & 0.512 & 0.482 & 0.547 & 0.512 & 0.434 & 0.547 & 0.484 \\
GPT-5.6-sol \citep{gpt56card2026}    & 0.555 & 0.679 & 0.611 & 0.589 & 0.681 & 0.632 & 0.558 & 0.679 & 0.612 \\
Claude Opus 4.8 \citep{opus48card2026} & 0.502 & 0.556 & 0.528 & 0.488 & 0.519 & 0.503 & 0.470 & 0.515 & 0.491 \\
Claude Sonnet 5 \citep{sonnet5card2026} & 0.536 & 0.602 & 0.567 & 0.579 & 0.566 & 0.572 & 0.510 & 0.570 & 0.539 \\
\midrule
\modelname (0.6B)            & 0.507 & 0.830 & 0.629 & 0.496 & 0.832 & 0.621 & 0.445 & 0.831 & 0.580 \\
\bottomrule
\end{tabular}
\end{table*}

%% file: tables/table_cdsize.tex
\begin{table*}[t]
\caption{Consistent detection by an entity's mention count and, for multi-mention entities, by turn
span: each cell is the fraction of that group's entities with \emph{every} mention covered.
Best per row in \textbf{bold}.}
\label{tab:cdsize}
\centering
\small
\setlength{\tabcolsep}{4pt}
\begin{tabular}{llccccccc|c}
\toprule
& & \multirow{2}{*}{Presidio} & \multirow{2}{*}{OpenAI PF} & \multirow{2}{*}{GLiNER2-PII}
& \multicolumn{2}{c}{GPT} & \multicolumn{2}{c}{Claude} & \multirow{2}{*}{\modelname} \\
\cmidrule(lr){6-7}\cmidrule(lr){8-9}
& & & & & 5.4 & 5.6-sol & Opus 4.8 & Sonnet 5 & \\
\midrule
\multirow{4}{*}{\rotatebox[origin=c]{90}{count}}
 & 1     & 0.655 & 0.474 & 0.641 & 0.829 & 0.788 & 0.840 & 0.829 & \textbf{0.917} \\
 & 2     & 0.671 & 0.397 & 0.356 & 0.413 & 0.689 & 0.333 & 0.449 & \textbf{0.873} \\
 & 3--5  & 0.630 & 0.263 & 0.258 & 0.216 & 0.504 & 0.216 & 0.252 & \textbf{0.796} \\
 & 6--10 & 0.627 & 0.182 & 0.073 & 0.082 & 0.464 & 0.045 & 0.118 & \textbf{0.691} \\
\midrule
\multirow{2}{*}{\rotatebox[origin=c]{90}{turn}}
 & cross-turn  & 0.648 & 0.296 & 0.292 & 0.275 & 0.551 & 0.222 & 0.305 & \textbf{0.776} \\
 & within-turn & 0.615 & 0.343 & 0.154 & 0.302 & 0.663 & 0.320 & 0.355 & \textbf{0.876} \\
\bottomrule
\end{tabular}
\end{table*}

%% file: tables/table_length.tex
\begin{table}[t]
\caption{\modelname character P/R/F1 by conversation length, read in a single 4{,}096-token window.
Docs: test documents per bucket; PII docs: those with gold spans; characters pooled per bucket.}
\label{tab:length}
\centering
\small
\begin{tabular}{lrrccc}
\toprule
Length & Docs & PII docs & P & R & F1 \\
\midrule
$<$1k     & 167     & 49  & 0.392 & 0.975 & 0.559 \\
1k--10k   & 1{,}292 & 505 & 0.516 & 0.955 & 0.670 \\
$\geq$10k & 463     & 243 & 0.507 & 0.687 & 0.583 \\
\midrule
Total     & 1{,}922 & 797 & 0.507 & 0.830 & 0.629 \\
\bottomrule
\end{tabular}
\end{table}

%% file: sections/conclusion.tex
\section{Conclusion}
\label{sec:conclusion}
In this work, we propose \dataname, a benchmark for PII detection in multi-turn LLM conversations, and \modelname, a compact 0.6B detector trained on such data. Experiments show that existing detectors either over-label PII-free conversations or miss some mentions of recurring identifiers; in contrast, \modelname more completely covers recurring identifiers and performs well on standard external benchmarks. We call for greater attention to PII detection in multi-turn LLM conversations.

%% file: sections/limitations.tex
\section*{Limitations}
\label{sec:limitations}

\dataname is designed as a controlled benchmark of PII detection in multi-turn
user--assistant conversations. Its conversations are synthetic reconstructions derived from the
structure of production assistant traffic. This construction supports public release with exact
span offsets and consistent mention chains, but it does not reproduce the source distribution
verbatim. The reported results should therefore be read as measurements of conversational PII
detection rather than estimates for any particular production workload.

The current release focuses on conversation text. Tool calls, inter-agent messages, and multimodal
inputs fall outside its scope, although they are natural extensions for studying how personal data
moves through agentic systems. Evaluation on conversations from additional assistants and domains
would also provide a broader test of transfer beyond the setting represented here.

Finally, each baseline is evaluated under a single inference configuration. Alternative
prompts, thresholds, context-window policies, or future model updates may change absolute scores.
The benchmark is thus best suited to comparing detector behavior within the evaluation setup used
here, rather than certifying production safety or regulatory compliance.

%% file: sections/ethics.tex
\section*{Ethical Considerations}
\dataname is built to strengthen protective systems: the intended use is evaluating and improving
PII detectors, not re-identifying anyone. The source conversations were accessed and processed
under the originating provider's terms of service and privacy policy; access was limited to
authorized project members under the provider's data-access controls, and the manual review of
audit-flagged documents took place under the same controls. No crowdworkers or external annotators
were involved: identifiers were labeled by prompted language models, and human review was limited
to the project team. We release only synthetic data: no source conversation is published,
every tagged identifier is a fabricated, leak-checked surrogate, and the known residual, untagged
personal names in some non-English text, is documented in the datasheet (Appendix~\ref{app:datasheet}).
The benchmark and the detector will be released under the MIT license. Scores on synthetic data
are proxies for, not guarantees of, behavior on real traffic, and the benchmark certifies neither
production safety nor regulatory compliance. Language models assisted with writing and engineering
in this project; the authors reviewed and verified all content and results.

%% file: sections/appendix.tex
\section{Datasheet for Datasets}
\label{app:datasheet}

Following \citet{gebru2021datasheets}, we answer the core questions inline; the release will
include a data card with the full questionnaire.
\paragraph{Motivation.} \dataname was created to evaluate PII detection \emph{in
multi-turn conversational context}, an axis missing from prior sentence- or record-level resources.
\paragraph{Composition.} Each instance is a synthetic multi-turn user--assistant conversation with
character-level identifier mentions over nine types, per-mention cluster ids, and a separate
sensitive-attribute layer. The corpus has 13{,}148 conversations across 13 languages
(train/validation/test $=$ 9{,}202 / 2{,}024 / 1{,}922), with per-record metadata.
Figure~\ref{fig:record} shows a complete released record.
\begin{figure*}[t]
\begin{lstlisting}
{"id": "conv_00042", "language": "en", "length_bucket": "<1k",
 "document_format": "unstructured", "clean_negative": false,
 "text": "U: I'm Dana Okoye, my number is +00 10 1111 0000.\nA: Thanks, Dana Okoye, I've noted the callback line.\nU: Please keep Dana Okoye on file.",
 "spans": [
   {"start": 7,   "end": 17,  "label": "private_person", "subtype": "full_name", "entity_id": 1, "entity_mentions": 3},
   {"start": 32,  "end": 48,  "label": "private_phone",  "subtype": "phone",     "entity_id": 2, "entity_mentions": 1},
   {"start": 61,  "end": 71,  "label": "private_person", "subtype": "full_name", "entity_id": 1, "entity_mentions": 3},
   {"start": 118, "end": 128, "label": "private_person", "subtype": "full_name", "entity_id": 1, "entity_mentions": 3}],
 "attribute_spans": []}
\end{lstlisting}
\caption{A complete released record (values fabricated).
Offsets index the \texttt{text} field; \texttt{entity\_id} links repeated mentions of one identifier, and
\texttt{entity\_mentions} is the cluster size used by consistent detection.}
\label{fig:record}
\end{figure*}

\input{tables/table_taxonomy}

\paragraph{Collection and annotation.} The corpus is derived from a deduplicated sample of real
user--assistant conversations from a production assistant; a ``turn'' is one delivered message.
To anonymize each conversation, multiple frontier LLMs locate the identifiers, following a written
guideline, and a rule-based pass groups repeated mentions of the same typed identifier into an
entity; sensitive attributes are marked separately. We inherit these labels as \dataname's gold.
The labeler prompt and guideline are
summarized in Appendix~\ref{app:prompts}, and the release will include them.

\section{Full Taxonomy and Crosswalk}
\label{app:taxonomy}
Table~\ref{tab:taxonomy} lists the nine identifier types with their sub-types and definitions, plus a crosswalk to standard regulatory categories; the full row-by-row mapping to OMB, NIST SP 800-122,
ISO/IEC 29100, and HIPAA will accompany the release. The label space subsumes those of the deployed
detectors we evaluate, so their outputs map directly into our types; the
residual \texttt{other\_pii} covers identifiers outside the eight concrete types.
The eight sensitive-attribute classes (health condition, religion, race/ethnicity, sexuality,
political view, disability, age, gender) form the separate attribute layer and are not part of the
identifier taxonomy.

\section{Detector: Additional Details}
\label{app:refmodel}

\paragraph{Training details.} We train with AdamW (learning rate $10^{-4}$, effective batch $256$)
on $8$ data-parallel GPUs for $3$ epochs, selecting the best checkpoint on held-out loss. The
training data and \dataname's sources share the same production traffic and matching conversation ids.

\paragraph{Frontier-LLM detector protocol.} The four frontier LLMs in Table~\ref{tab:cd} are
prompted zero-shot, with no few-shot examples. A system message states the task and the nine
identifier types and asks for a JSON list of \{\texttt{text}, \texttt{type}\} objects whose
\texttt{text} is copied verbatim from the input (Appendix~\ref{app:prompts}). Documents are read
in $8{,}000$-character windows with $400$-character overlap; each returned string is mapped back
to character offsets by exact substring match, and offsets are merged and de-duplicated across
windows.

\section{Extended Results}
\label{app:results}

\paragraph{Full detection results.} Table~\ref{tab:agg} completes the entity-level view of
\S\ref{sec:crossturn} for all twelve detectors. \modelname leads on all three coverage columns
(CD $0.853$, CD-m $0.794$, xCD $0.776$). Its FP$_0$ of $0.385$ is comparable to the frontier LLMs
($0.231$--$0.386$) and well below every public specialized baseline ($0.557$--$0.972$). Coverage and
false positives must be read together, since a detector can raise CD by flagging nearly every
document, as Presidio and the OpenMed models do.
\begin{table}[t]
\caption{Entity-level results on the test split: consistent detection (CD), multi-mention CD
(CD-m), cross-turn CD (xCD), and PII-free false-positive rate FP$_0$ (lower is better). Best per
column in \textbf{bold}.}
\label{tab:agg}
\centering
\small
\setlength{\tabcolsep}{4pt}
\begin{tabular}{@{}lcccc@{}}
\toprule
Detector & CD & CD-m & xCD & FP$_0$ \\
\midrule
Presidio        & 0.649 & 0.642 & 0.648 & 0.972 \\
OpenAI PF       & 0.386 & 0.304 & 0.296 & 0.557 \\
GLiNER2-PII     & 0.448 & 0.268 & 0.292 & 0.862 \\
GLiNER-PII      & 0.472 & 0.349 & 0.334 & 0.901 \\
Piiranha        & 0.128 & 0.090 & 0.097 & 0.685 \\
OpenMed 44M     & 0.724 & 0.745 & 0.761 & 0.936 \\
OpenMed 434M    & 0.709 & 0.743 & 0.759 & 0.935 \\
GPT-5.4         & 0.545 & 0.279 & 0.275 & 0.321 \\
GPT-5.6-sol     & 0.675 & 0.570 & 0.551 & \textbf{0.231} \\
Claude Opus 4.8 & 0.530 & 0.239 & 0.222 & 0.386 \\
Claude Sonnet 5 & 0.563 & 0.314 & 0.305 & 0.276 \\
\midrule
\modelname      & \textbf{0.853} & \textbf{0.794} & \textbf{0.776} & 0.385 \\
\bottomrule
\end{tabular}
\end{table}

\paragraph{Inference policy for long documents.} Table~\ref{tab:policy} decodes the same
\modelname checkpoint under three long-input policies. Relative to the single window, sliding
windows raise recall on documents at or above 10k characters from $0.687$ to $0.975$ and CD-m from
$0.794$ to $0.954$, while character precision drops from $0.507$ to $0.493$; chunked decoding lies
between the two.
\begin{table}[!t]
\caption{\modelname under three long-input decoding policies: a single window (trunc),
non-overlapping windows (chunk), and 50\%-overlap sliding windows (slide). R$_{<10k}$ and
R$_{\geq 10k}$ split character recall at 10k characters.}
\label{tab:policy}
\centering
\small
\setlength{\tabcolsep}{2.6pt}
\begin{tabular}{@{}lccccccc@{}}
\toprule
Policy & P & R & F1 & CD & CD-m & R$_{<10k}$ & R$_{\geq 10k}$ \\
\midrule
trunc  & 0.507 & 0.830 & 0.629 & 0.853 & 0.794 & 0.956 & 0.687 \\
chunk  & 0.502 & 0.944 & 0.656 & 0.916 & 0.894 & 0.956 & 0.931 \\
slide  & 0.493 & 0.965 & 0.653 & 0.954 & 0.954 & 0.957 & 0.975 \\
\bottomrule
\end{tabular}
\end{table}

\section{Prompts}
\label{app:prompts}
The four LLM prompts of this work are shown below.

\begin{tcolorbox}[fancyprompt, title={\faHighlighter\ Gold Labeler Prompt (\S\ref{sec:construction})}]
\textbf{\color{PPLXDeep}Task.} You are an expert PII labeler producing gold labels for \ldots\
conversations. Your labels will train a downstream PII-masking model, so they must reflect
industry-standard definitions of Personally Identifiable Information\,\ldots
\par\medskip
\textbf{\color{PPLXDeep}Test.} Would a knowledgeable reader, given this string and the
surrounding context, be able to identify or contact a specific person? If the string instead
identifies a publicly-listed organization, a public-figure persona, or a fictional vignette
persona, the answer is no\,\ldots
\par\medskip
\textbf{\color{PPLXDeep}Traps.} Even when the string looks exactly like a name, address, phone,
URL, or date, emit nothing if the context matches a trap: a named officer or staff member in an
organizational role, reserved or documentation values, citation or tool tokens\,\ldots
\par\medskip
\textbf{\color{PPLXDeep}Output.} A single JSON object with two fields: a conversation-level
sensitivity head and the token-level PII span list of \{\texttt{label}, \texttt{text}\}
objects\,\ldots
\end{tcolorbox}

\begin{tcolorbox}[fancyprompt, title={\faPenNib\ Paraphrase Prompt (Figure~\ref{fig:pipeline}, step~3)}]
\textbf{\color{PPLXDeep}Task.} Rewrite a snippet from a real user--assistant chat so it is not
verbatim the original, while preserving its meaning, intent, tone, language, and Markdown
structure (headings, lists, tables, code fences\,\ldots).
\par\medskip
\textbf{\color{PPLXDeep}Rule 1: placeholders.} The text may contain special placeholder
characters; each stands for a redacted value. Reproduce every placeholder exactly once, in the
natural grammatical position where its value belongs; never delete, duplicate, translate\,\ldots
\par\medskip
\textbf{\color{PPLXDeep}Rule 2: de-identify context.} Any real-world specific that is not a
placeholder (company, product, or brand names; unusual place names; distinctive verbatim
phrases\,\ldots) must become a generic or plausibly invented equivalent.
\par\medskip
\textbf{\color{PPLXDeep}Output.} Roughly the same length and the same language as the input;
only the rewritten snippet, no preamble\,\ldots
\end{tcolorbox}

\begin{tcolorbox}[fancyprompt, title={\faSearch\ Second-LLM Audit Prompt (\S\ref{sec:verify})}]
\textbf{\color{PPLXDeep}Task.} Audit synthetic-conversation text for a PII benchmark; all names,
contacts, and IDs are supposed to be fabricated\,\ldots
\par\medskip
\textbf{\color{PPLXDeep}Report.} Only strings that look like a real, specific individual's
genuine identifying data that fabrication might have missed (an intact real email, phone, or
handle; a real public figure named as the actual user\,\ldots).
\par\medskip
\textbf{\color{PPLXDeep}Ignore.} Generic names, places, and dates; clearly synthetic values.
\par\medskip
\textbf{\color{PPLXDeep}Output.} JSON \texttt{\{"suspect": bool,
"items": [\{"text", "why"\}\,\ldots]\}}.
\end{tcolorbox}

\begin{tcolorbox}[fancyprompt, title={\faLightbulb\ Frontier-LLM Detector Prompt (\S\ref{sec:experiments})}]
\textbf{\color{PPLXDeep}Task.} You are a precise PII detector operating on text from a multi-turn
assistant conversation. Extract every span that is personal information identifying a private
individual\,\ldots
\par\medskip
\textbf{\color{PPLXDeep}Rule.} Mark a span only when, in this conversational context, it helps
identify a private person; do not mark public figures, organizations, fictional or placeholder
values\,\ldots\ A date counts only when it identifies a person\,\ldots
\par\medskip
\textbf{\color{PPLXDeep}Output.} A JSON list of \{\texttt{text}, \texttt{type}\} objects over the
nine identifier types, with \texttt{text} copied verbatim from the input\,\ldots
\end{tcolorbox}

%% file: tables/table_taxonomy.tex
\begin{table*}[t]
\centering
\small
\caption{The nine \dataname identifier types: sub-types, one-line definition (a span is labeled
only when it identifies a \emph{private} person; placeholders and public entities are excluded), and
regulatory crosswalk.}
\label{tab:taxonomy}
\begin{tabular}{l p{5.8cm} p{5.8cm}}
\toprule
Type & Sub-types & Definition / crosswalk \\
\midrule
\texttt{private\_person}  & full name, name part, social handle/username & name of a private person; OMB/NIST direct, HIPAA name \\
\texttt{private\_email}   & email address                 & personal email; NIST, HIPAA electronic mail \\
\texttt{private\_phone}   & phone/fax number              & phone of a private person; NIST, HIPAA \\
\texttt{private\_address} & postal address, precise geo   & location of a private person; HIPAA geographic \\
\texttt{private\_url}     & profile URL, IP address       & URL/IP for a private audience; NIST online identifier \\
\texttt{private\_date}    & date of birth, identifying date & datetime identifying a person; HIPAA dates \\
\texttt{account\_number}  & national ID, SSN, IBAN, credit card, sort code, id number & account/record identifier; NIST, HIPAA account no. \\
\texttt{secret}           & password/API key, CVV/PIN     & credential; NIST authentication data \\
\texttt{other\_pii}       & residual annotator-marked identifier & person-identifying but outside the eight types \\
\bottomrule
\end{tabular}
\end{table*}